%% file: main.tex
\documentclass[letterpaper, 10 pt, conference]{IEEEconf}

\IEEEoverridecommandlockouts                              

\usepackage{graphics} 
\usepackage{mathtools}
\usepackage{times} 
\usepackage{amsmath} 
\usepackage{amssymb}  
\usepackage{xcolor}
\usepackage[utf8]{inputenc}
\usepackage{comment}
\usepackage{algorithm,algorithmic}
\usepackage{amsthm}
\usepackage{balance}
\usepackage{hyperref}
\DeclareMathAlphabet\mathbfcal{OMS}{cmsy}{b}{n}

\usepackage{cancel}
\theoremstyle{plain}
\newtheorem{definition}{Definition}
\newtheorem{remark}{Remark}

\newcommand{\q}{\boldsymbol{q}}
\newcommand{\R}{\mathbb{R}}
\newcommand{\Poin}{\mathcal{P}}
\newcommand{\x}{\boldsymbol{x}}
\newcommand{\e}{\boldsymbol{e}}
\newcommand{\Y}{\mathbfcal{Y}}
\newcommand{\h}{\boldsymbol{h}}
\newcommand{\bxi}{\boldsymbol{\xi}}

\newcommand{\ti}{\text{i}}
\newcommand{\tOA}{\text{OA}}
\newcommand{\tUA}{\text{UA}}
\newcommand{\tFA}{\text{FA}}

\DeclareMathAlphabet{\mathcal}{OMS}{cmsy}{m}{n}

\title{\LARGE \bf
Multi-Domain Walking with Reduced-Order Models of Locomotion}

\author{Min Dai, Jaemin Lee and Aaron D. Ames
\thanks{ The authors are with the Department of Mechanical and Civil Engineering, California Institute of Technology, Pasadena, CA 91125 USA.
        {\tt\small \{mdai,  jaemin87, ames\}@caltech.edu}. This work is supported by NSF NRI award 1924526 and NSF CMMI award 1923239. }%
}

\begin{document}

\maketitle
\thispagestyle{empty}
\pagestyle{empty}

\begin{abstract}
Drawing inspiration from human multi-domain walking, this work presents a novel reduced-order model based framework for realizing multi-domain robotic walking. 
At the core of our approach is the viewpoint that human walking can be represented by a hybrid dynamical system, with continuous phases that are fully-actuated, under-actuated, and over-actuated and discrete changes in actuation type occurring with changes in contact.  Leveraging this perspective, we synthesize a multi-domain linear inverted pendulum (MLIP) model of locomotion.  
Utilizing the step-to-step dynamics of the MLIP model, we successfully demonstrate multi-domain walking behaviors on the bipedal robot Cassie---a high degree of freedom 3D bipedal robot. Thus, we show the ability to bridge the gap between multi-domain reduced order models and full-order multi-contact locomotion.  Additionally, our results showcase the ability of the proposed method to achieve versatile speed-tracking performance and robust push recovery behaviors. 
\end{abstract}

\input{intro}

\input{prelim}

\input{romodel}

\input{fullmodel}
\input{results}



\bibliographystyle{IEEEtran}
\balance

\bibliography{references,misc}

\end{document}

%% file: intro.tex
\section{Introduction}

\begin{figure}[t]
    \centering
    \includegraphics[width = 1 \linewidth ,height=7cm]{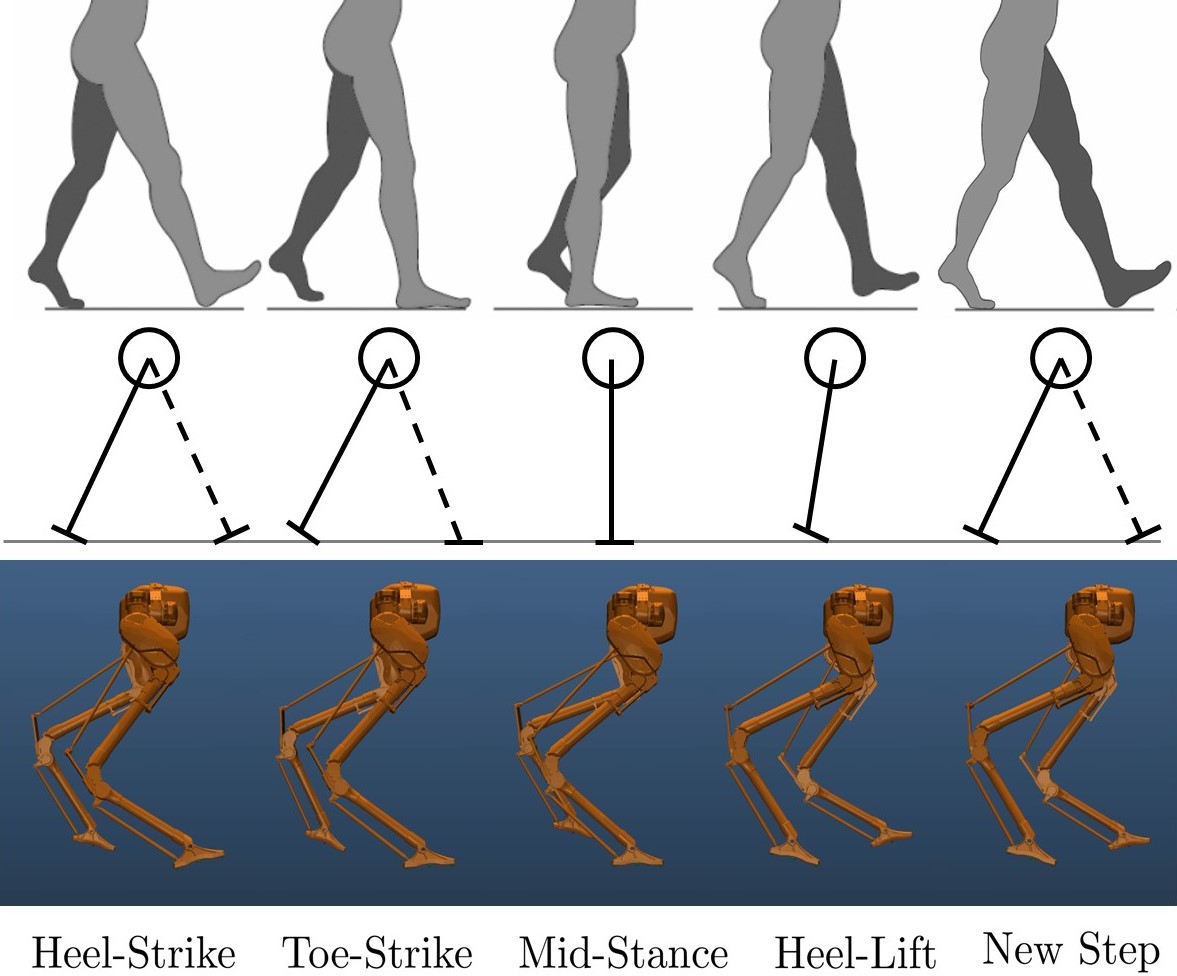}
    \vspace{-6mm}
    \caption{A complete gait cycle: (top) Human multi-domain walking. (center) MLIP walking inspired by human walking. (bottom) Cassie's multi-domain walking stabilized through the MLIP model.}
    \vspace{-4mm}
    \label{fig::gaitcycle_for_all}
\end{figure}

The agility and versatility displayed in human locomotion have long served as an inspiration for the study of robotic bipedal locomotion. For humans, walking involves a sequence of distinct gait phases as shown in Fig. \ref{fig::gaitcycle_for_all}. In the context of forward walking, these phases encompass the swing foot's heel strike, toe strike, the transition of weight from the new stance foot's heel to toe, and the subsequent heel lift and ankle push-off of the stance foot \cite{ames2011human}. 
In contrast, walking robots typically rely on flat-footed gaits. This preference often arises from mathematical convenience: a desire for a feedback-linearizable fully-actuated system \cite{kajita_3d_2001, humanoid_reference} or a direct application of point-foot walking methods on robots with conventional feet \cite{gong_zero_2022, xiong_3-d_2022}. Nevertheless, multi-domain gait presents compelling biomechanical advantages: the heel strike effectively dampens impact forces and ankle push-off is remarkably energy-efficient. These advantages have been substantiated in robotic applications as in \cite{kim_once-per-step_2017,reher2020algorithmic}. Furthermore, compared to flat-foot gait, the capacity to raise the heel permits longer strides within the same joint constraints, resulting in faster walking speeds \cite{sellaouti_faster_2006}.



Researchers have explored methods for realizing multi-domain walking on robotic platforms due to its advantages. Full-model based methods \cite{chevallereau_stable_2008, zhao_multi-contact_2017, chao_step_2017,reher2020algorithmic} employ multi-domain trajectory optimization within the Hybrid Zero Dynamics (HZD) framework. They entail solving a challenging nonlinear optimization problem, which can be computationally demanding and prone to convergence issues for obtaining a single periodic orbit. Additionally, these methods necessitate offline trajectory generation for different periodic orbits associated with different speeds and contact sequences. Furthermore, they are sensitive to model discrepancies and require a heuristic foot placement regulator \cite{reher_dynamic_2021} for stabilization---often synthesized from reduced-order models, i.e., the ``Raibert controller'' \cite{raibert1986legged}.


In the realm of prosthesis legs and feet, multi-domain walking is often utilized but with ``model-free'' controllers. These controllers typically consist of combinations of low-gain impedance control \cite{sup_design_2008, fey_controlling_2014} and torque-based control, which is determined either using predefined parameters \cite{thatte_robust_2019} or through a biomechanically inspired model \cite{eilenberg_control_2010}. This flexible controller design allows for expert tuning to tailor parameters to individual users under different scenarios, but this tuning is time-consuming. In addition, stability is not a primary concern in this domain, as it is assumed that the human user can stabilize themselves through stepping. Some approaches incorporate the human into the HZD gait generation loop \cite{gehlhar_review_2023}, but they encounter similar to those previously mentioned. 


In this paper, we present a novel framework that enables multi-domain walking, including: heel-to-toe, toe-to-heel, and flat-footed behaviors. We represent these different phases of locomotion through a hybrid dynamical system model.  
Leveraging this, and as illustrated in Fig. \ref{fig::gaitcycle_for_all}, our approach begins with the introduction of a reduced-order model, the \emph{multi-domain linear inverted pendulum (MLIP)}, specifically designed to capture the intricate weight-shifting dynamics inherent in multi-domain walking. Subsequently, we develop a controller that stabilizes the Poincar\'{e} map associated with the hybrid dynamics of the MILP, i.e., ensures the stability of the step-to-step dynamics.  Via the construction of outputs for the full-order system, we synthesize a feedback controller that realizes multi-domain walking on the bipedal robot Cassie. This results in a remarkably human-like walking gait, adhering to the same gait cycle time distribution observed in human walking. 
Leveraging the effective ankle push-off during heel-toe-toe walking, we attain an impressive walking speed of 2.15 m/s on Cassie, which cannot be realized using a flat-footed gait due to inherent physical joint limits. Notably, our method surpasses existing approaches by offering versatile walking behaviors adaptable to diverse gait parameters and commanded speeds, all while guaranteeing stability and eliminating the need for offline optimization. Furthermore, our framework demonstrates robustness against external disturbances. A collection of the resulting walking behaviors is available in the accompanying video\footnote{\url{https://youtu.be/8u5ZiWe_qlw}}.


The rest of the paper is structured as follows: Section \ref{sec::hybrid_robot} introduces the hybrid control problem of multi-domain walking. Sections \ref{sec::ro_model} and \ref{sec::full_model} detail the proposed MLIP model and its integration into the full robot model, respectively.  We then present the results and evaluate the performance under different circumstances in section \ref{sec::results}. Finally, the paper concludes with Section \ref{sec::conclusion}.

%% file: prelim.tex
\section{Hybrid Dynamics of Bipedal Robots}\label{sec::hybrid_robot}

Bipedal walking is represented by a hybrid control system \cite{reher_dynamic_2021,grizzle_models_2014} defined as the tuple $\mathcal{HC} = (\Gamma, \mathcal{D}, \mathcal{S}, \Delta, \mathcal{FG})$. Each component is explained as follows:
\begin{itemize}
    \item $\Gamma = (V,E)$ is a directed circle graph with a set of vertices $V = \{ v_\text{i}\}_{\ti \in I}$ and a set of directed edges $E = \{   v_\ti \hspace{-.1cm}\rightarrow \hspace{-.1cm} v_\text{j}   \}_{\ti,\text{j} \in I}$. $I$ is an indexed set of domains that we will illustrate soon.
    \item $\mathcal{D} = \{\mathcal{D}_v\}_{v\in V}$ is a set of domains of admissibility.
    \item $\mathcal{S} = \{ \mathcal{S}_e\}_{e \in E}$ is a set of guards. 
    \item $\Delta = \{ \Delta_e\}_{e \in E}$ is a set of reset maps
    \item $\mathcal{FG} = \{f_v, g_v\}_{v\in V}$ is the continuous control system, which is a set of vector fields on the state manifolds.
\end{itemize}

In any domain, the robot's continuous dynamics can be obtained from the Euler-Lagrangian equations:
\begin{align}
    &D(\q)\ddot{\q} + H(\q,\dot{\q}) = B \boldsymbol{\tau} + J_\ti(\q)^T \boldsymbol{f}_\ti ,\label{eq::eom} \\ 
    &J_\ti(\q)\ddot{\q} + \dot{J}_\ti(\q,\dot{\q})\dot{\q} = 0, \label{eq::hol} 
\end{align}
where $\boldsymbol{q}\in Q$ is a set of generalized coordinates in the $n$-dimensional configuration space $Q$, $D(\q)\in \R^{n \times n}$, $H(\q,\dot{\q}) \in \R^{n}$, $B\in \R^{n\times m}$ are the inertia matrix, the collection of centrifugal, Coriolis and gravitational forces, and the actuation matrix, respectively. Additionally, $\boldsymbol{\tau}\in U \subseteq \R^m$ stands for input torque,  $J_\ti(\q)\in\R^{n\times h_\ti}$ is the domain-specific Jacobian matrix related to contact constraints, and $\boldsymbol{f}_\ti\in\R^h_\ti$ represents the corresponding constraint wrench.

Discrete impacts are assumed to be instantaneous and plastic, with solution derivations detailed in \cite{grizzle_models_2014}. Denoting $\x = [\q^T, \dot{\q}^T]^T \in \mathcal{TQ}$, the equation of motion for the hybrid system is as follows:
\begin{align}
    \begin{cases} \dot{\x} &= f_v(\x) + g_v(\x) \boldsymbol{\tau}  \quad \x \in \mathcal{D}_v \setminus  \mathcal{S}_e\\
    \x^+ &= \Delta_e( \x^-) \hspace{1.4cm} \x^-\in \mathcal{S}_e  \end{cases},
    \label{eq::HZ}
\end{align}
for all $v \in V $ and corresponding $e \in E$.

\begin{figure}[t]
    \centering
    \includegraphics[width = 1 \linewidth]{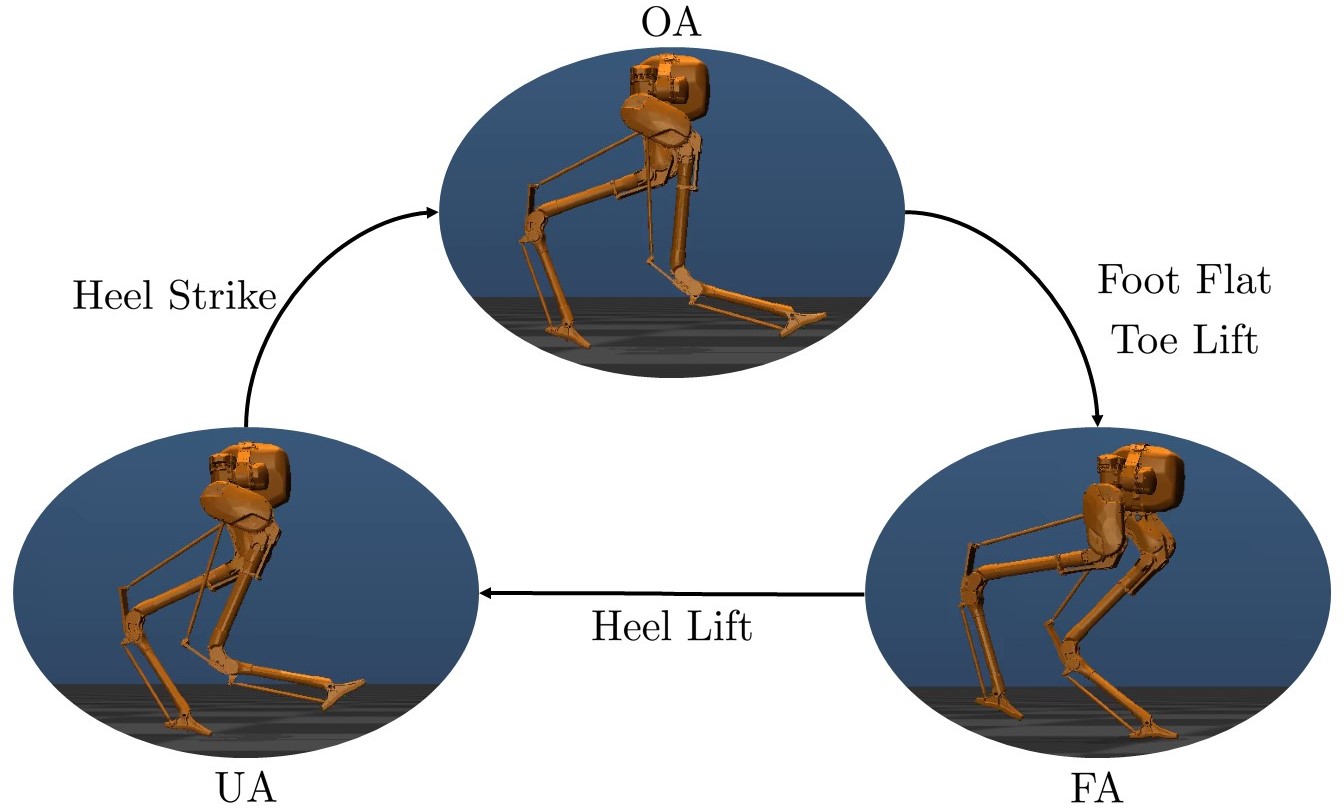}
    \vspace{-6mm}
    \caption{Hybrid system directed graph for heel-to-toe walking}
    \vspace{-4mm}
    \label{fig::domain_graph}
\end{figure}

Inspired by human heel-to-toe walking, our hybrid system model incorporates three domains: fully-actuated (FA), under-actuated (UA) and over-actuated (OA). This yields $V = { v_\tFA, v_\tUA, v_\tOA }$ and $ E ={ v_\tFA \hspace{-.1cm}\rightarrow \hspace{-.1cm} v_\tUA , v_\tUA \hspace{-.1cm}\rightarrow \hspace{-.1cm} v_\tOA , v_\tOA \hspace{-.1cm}\rightarrow \hspace{-.1cm}v_\tFA }$, as illustrated in Fig. \ref{fig::domain_graph}.

\begin{definition}
We define the domains for a hybrid dynamical system of bipedal robots as follows:\\
    (FA) Fully-actuated domain: This describes the single support phase where the stance foot has full ground contact. \\
    (UA) Under-actuated domain: This characterizes the single support phase where only the stance toe has ground contact. \\
    (OA) Over-actuated domain: This corresponds to the double support phase, with both the back leg toe and the front leg heel in contact with the ground.
\end{definition}

\begin{definition}\label{def::guard}
Guards are defined to represent the switching condition from one domain to the next following defined edges. More specifically, the transition from the FA to UA domain occurs when the stance heel lifts off the ground. The guard is given by:
\begin{align*}
    \mathcal{S}_{\tFA \rightarrow \tUA} &\coloneqq \{ (\x,\boldsymbol{\tau}) \in \mathcal{TQ} \times U :  f^z_\text{st,heel}(\x,\boldsymbol{\tau}) = 0  \}.
\end{align*}
In addition, the UA to OA domain transition takes place when the swing heel strikes the ground, as indicated by:
\begin{align*}
    \mathcal{S}_{\tUA \rightarrow \tOA} &\coloneqq \{ \x \in \mathcal{TQ}:  z_\text{sw,heel}(\x) = 0  \}.
\end{align*}
Lastly, transitioning from the OA to FA domain occurs when the swing toe contacts the ground, and the previous stance toe lifts off, providing the following guard condition:
\begin{align*}
    \mathcal{S}_{\tOA \rightarrow \tFA} \coloneqq \{ (\x,\boldsymbol{\tau}) \in \mathcal{TQ} \times U : &  z_\text{sw,toe}(\x) = 0,\\
    &f^z_\text{st,toe}(\x,\boldsymbol{\tau}) = 0  \}.
\end{align*}
\end{definition}




\begin{remark}
   To conserve space, we have exclusively presented the heel-to-toe walking mode. One can equivalently define $\mathcal{HC}$ models for toe-to-heel walking and flat-footed walking. In flat-footed walking, the UA domain corresponds to the deactivation of the stance ankle torque. 
\end{remark}


%% file: romodel.tex
\section{Step-to-Step Dynamics and Stabilization for Multi-Domain LIP Model}\label{sec::ro_model}

In this section, we first propose a reduced-order model, termed the multi-domain linear inverted pendulum (MLIP) model, which is a variant of the canonical linear inverted pendulum (LIP) model \cite{kajita_3d_2001} that can describe multi-domain walking. This extension involves incorporating the position of the zero-moment point (ZMP)  \cite{vukobratovic_zero-moment_2004} as an additional state variable. After characterizing its step-to-step (S2S) dynamics, we apply a linear controller to stabilize the error dynamics, considering the discrepancy between the actual dynamics of the robot and the reduced-order model.

\begin{figure}[t]
    \centering
    \includegraphics[width = 1 \linewidth]{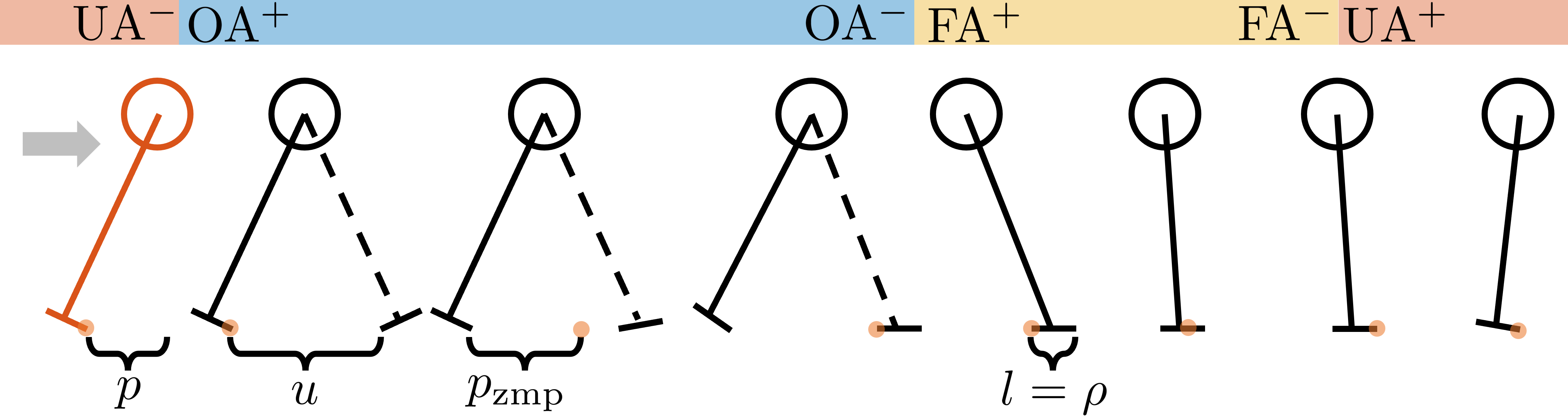}
    \vspace{-6mm}
    \caption{MLIP gait cycle for heel-to-toe walking. Note that as the MLIP model has massless legs and feet, the foot angles do not impact dynamics. The stance and swing foot pitch angles shown are up to the user's choice. The feet drawn are straight lines, but curved feet with an arc length $\rho$ would result in the same dynamics.}
    \label{fig::lip_h2t_defintion}
\end{figure}

\begin{figure}[t]
    \centering
    \includegraphics[width = 1 \linewidth]{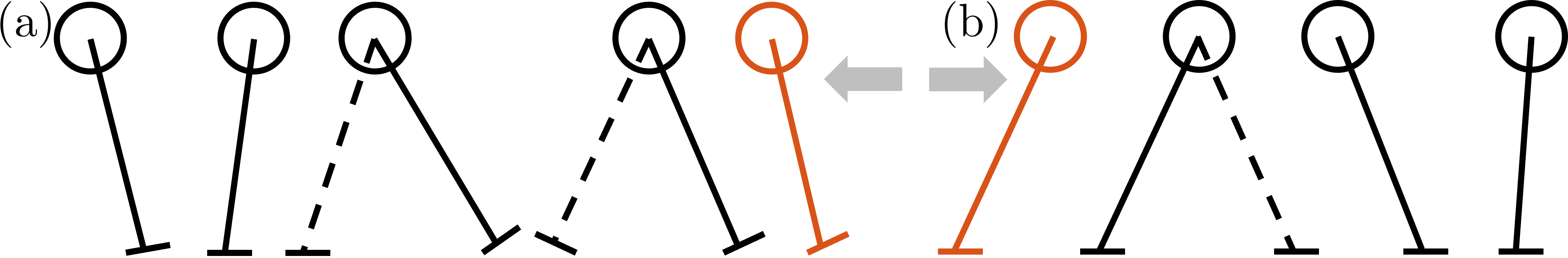}
    \caption{MLIP gait cycle for (a) toe-to-heel and (b) flat-footed walking with arrows indicating walking direction. }
    \vspace{-4mm}
    \label{fig::lip_t2h_flat_defintion}
\end{figure}

\subsection{MLIP Model and Step-to-step Dynamics}
As shown in Fig. \ref{fig::lip_h2t_defintion} and \ref{fig::lip_t2h_flat_defintion}, the MLIP model includes a point mass and two massless telescopic legs. It also has a constant center of mass (CoM) height $z_0$ relative to the stance pivot as in the LIP model. What sets it apart from the conventional LIP model is the inclusion of a pair of feet with a known arc length denoted as $\rho$, which can be calculated from foot curvature. In accordance with the bipedal locomotion domains outlined in Sec. \ref{sec::hybrid_robot}, the MLIP model encompasses the UA, OA, and FA domains. The continuous dynamics of the MLIP model in all domains are governed by the following linear equations:
\begin{align}
\frac{d}{dt}
    \underbrace{\begin{bmatrix}
        p \\ L \\ p_\text{zmp}
    \end{bmatrix}}_{\bxi} =
    \underbrace{\begin{bmatrix}
        0 & \frac{1}{z}  & 0\\
        g & 0 &-g \\ 
        0&0&0
    \end{bmatrix} }_{A_\text{ct}}
    \begin{bmatrix}
        p \\ L \\ p_\text{zmp}
    \end{bmatrix}+
    \underbrace{\begin{bmatrix}
        0 \\ 0 \\ 1
    \end{bmatrix}}_{B_\text{ct}} \dot{p}_\text{zmp},
\end{align}
where $p$, $L$, and $p_\text{zmp}$ represent the horizontal CoM position, mass-normalized centroidal angular momentum \cite{orin_centroidal_2013}, and horizontal ZMP position, all defined relative to the stance pivot. The stance pivot we refer to is the stance pivot at the UA phase thus it is dependent on walking mode. When walking in a heel-to-toe manner, the stance pivot is positioned at the toe, while in toe-to-heel walking, it is located at the heel. For flat-footed walking, the stance pivot can be picked anywhere from heel to toe. In this context, we choose the pivot point to be directly under the ankle. Given that the dynamics are linear, the end-of-domain states $\bxi_{\ti}^{-}$ has closed-form solution given by:
\begin{align} \label{eq::MLIP_ct_raw}
    \bxi_\ti^- = \underbrace{e^{A_\text{ct} T_\text{i}}}_{A_\text{i}} \bxi_\text{i}^+ + \int_0^{T_\text{i}} e^{A_\text{ct} (T_\text{i}-t)} B_\text{ct} \dot{p}_\text{zmp, i}(t) dt,
\end{align}
where superscripts $(\cdot)^{+ / -}$ indicate the beginning and end of each domain, respectively, and $T_{\ti}$ represents the time duration of the i-th domain. Unlike the guards defined in Definition \ref{def::guard}, the transitions between domains in the MLIP model are purely time-based, as the legs are effectively virtual. To determine $\dot{p}_\text{zmp, i}$ in each domain, we draw inspiration from human walking data \cite{danilov_zmp_2016}, as elaborated below.

\noindent \textbf{FA: } We denote the distance that the ZMP travels during the fully-actuated phase as $l$. In the context of heel-to-toe walking, this distance corresponds to $\rho$. For toe-to-heel and flat-footed walking, the values are $-\rho$ and 0, respectively. 
\begin{align}
    \dot{p}_\text{zmp, FA}(t) = \frac{l}{T_\text{FA}}.
\end{align}

\noindent \textbf{UA: } During the under-actuated phase, the ZMP is considered to be fixed at the stance pivot. Thus, $\dot{p}_\text{zmp, UA}(t) = 0$.

\noindent \textbf{OA: } During the over-actuated phase, the ZMP shifts from the previous stance pivot to the new stance leg, thus traveling the step length denoted as $u$ within the step duration $T_\text{OA}$:
\begin{align}
    \dot{p}_\text{zmp, OA}(t) = \frac{u}{T_\text{OA}}.
\end{align}
As depicted in Fig. \ref{fig::lip_h2t_defintion}, in the context of heel-to-toe walking, the parameter $u$ indicates the distance between the stance toe and the swing heel. However, in the case of toe-to-heel and flat-footed walking, $u$ is adapted to denote the distance between the stance heel and the swing toe, as well as the separation between the stance under the ankle and the swing under the ankle.

Given $\dot{p}_\text{zmp, i}$ is defined to be independent of time for all domains, we further simplified the above dynamics \eqref{eq::MLIP_ct_raw} :
\begin{align*}
    \int_0^{T_\text{i}} e^{A_\text{ct} (T_\text{i}-t)} B_\text{ct} \dot{p}_\text{zmp, i}(t) dt &= \int_0^{T_\text{i}} e^{A_\text{ct} (T_\text{i}-t)} dt B_\text{ct} \dot{p}_\text{zmp, i}\\
    &=\underbrace{\int_0^{T_\text{i}} e^{A_\text{ct} (T_\text{i}-t)} dt B_\text{ct} \frac{1}{T_\text{i}}}_{B_\text{i}}d_\text{i}.
\end{align*}
As a result, we can express the solution to the continuous time dynamics for $i$-th domain as follows:
\begin{align}\label{eq::MLIP_CT}
    \bxi_\ti^- = A_\ti \bxi_\ti^+ + B_\ti d_\ti
    \tag{MLIP-CT}
\end{align}
where $d_\text{OA} = u$, $d_\text{FA}=l$, and $d_\text{UA}=0$.


With continuous phase dynamics defined, we need to specify the impact dynamics. For the model with massless legs, discrete state jumps resulting from impact do not occur. Instead, the impact equation characterizes the effects of switching between the stance and swing legs, which is defined to happen at the transition from OA to FA domain. Thus, the impact equations are given by:
\begin{align} \label{eq::MLIP_DT}
\begin{cases}
   \bxi_\text{OA}^+ &= \bxi_\text{UA}^-\\
    \bxi_\text{FA}^+ &= \bxi_\text{OA}^- + B_\Delta  u + C_\Delta\\
    \bxi_\text{UA}^+ &= \bxi_\text{FA}^-   
\end{cases}    ,
\tag{MLIP-DT}
\end{align}
where $B_\Delta = \begin{bmatrix}
    -1 & 0 & -1
\end{bmatrix}^T$ and $C_\Delta = \begin{bmatrix}
    -l & 0 & -l
\end{bmatrix}^T$.


We are interested in understanding the S2S dynamics for stabilizing the system. Consider the pre-impact state at the UA phase as $\bxi[k]$ in the $k$-th step, which is shown in red in Fig. \ref{fig::lip_h2t_defintion} and \ref{fig::lip_t2h_flat_defintion}, a complete step evolution follows this sequence: $\bxi[k] \rightarrow \bxi_\text{OA}^+ \rightarrow \bxi_\text{OA}^- \rightarrow \bxi_\text{FA}^+ \rightarrow \bxi_\text{FA}^- \rightarrow \bxi_\text{UA}^+ \rightarrow \bxi[k+1]$. Consequently, the S2S dynamics of the MLIP model can be described using Eq. \eqref{eq::MLIP_DT} and \eqref{eq::MLIP_CT}:
\begin{align}\label{eq::MLIP_S2S_3}
    \bxi_{k+1} = A_{\bxi} \bxi_k + B_{\bxi} u_k + C_{\bxi}
\end{align}
where the detailed matrices are obtained as follows:
\begin{subequations}
\begin{align}
    A_{\bxi} &=  A_\tUA A_\tFA A_\tOA, \\
    B_{\bxi} &=  A_\tUA A_\tFA (B_\tOA + B_\Delta), \\
    C_{\bxi} &=  A_\tUA A_\tFA (B_\tFA l + C_\Delta) .
\end{align}
\end{subequations}
Given that the Poincaré section is defined at the end of the UA phase, it follows that $p_\text{zmp,k} = 0$ for all $k \in \mathbb {N}$. Additionally, it can be confirmed that $A_{\bxi}(3,3) = 1$, $B_{\bxi}(3) = 0$, and $C_{\bxi}(3) = 0$. Therefore, we use $\x^M \in \R^2$ to represent the horizontal CoM states, specifically $\x^M = [p , L]^T$, such that $\bxi = [\x^M; 0]$. Using the two-dimensional states, we can express the S2S dynamics of the MLIP as follows:
\begin{align}\label{eq::MLIP_S2S}
    \x_{k+1}^\text{M} = A^\text{M} \x_k^\text{M} + B^\text{M} u_k + C^\text{M} , \tag{MLIP-S2S}
\end{align}
where $A^\text{M} = A_{\bxi}(1:2,1:2)$, $B^\text{M} = B_{\bxi}(1:2)$, and $C^\text{M} = C_{\bxi}(1:2)$.

\begin{remark}
It might appear that the MLIP model requires non-trivial FA, UA, and OA phases to obtain well-defined S2S dynamics, given the presence of $\frac{1}{T_\ti}$ in the definition of $B_\ti$. However, it's important to note that the integral from 0 to $T_\ti$ always results in zero when $T_\text{i} =0$. In cases of $T_\text{OA} =0$, we also need to set $B_\Delta = \begin{bmatrix}
-1 & 0 & 0
\end{bmatrix}^T$ to immediately transfer the zero moment point to the new stance foot.
\end{remark}

\begin{remark}
    The second coordinate $L$ can be replaced by linear velocity, as done in \cite{kajita_3d_2001, xiong_3-d_2022}. However, we have chosen to use angular momentum about stance pivot for higher data quality, given a Kalman filter for angular momentum can be easily set up as demonstrated in \cite{gong_angular_2021}. Replacing the second coordinate will result in a modified $A_\text{ct}$, while the linear structure of the step-to-step dynamics remains unchanged.
\end{remark}

\begin{remark}
    One can retrieve the Hybrid-LIP model in \cite{xiong_3-d_2022} by using the CoM linear velocity as the second coordinate and setting $l=0$ and $T_\text{FA} = T_\text{OA} = 0$. However, there are different assumptions regarding the motion of the ZMP in the OA phase, which is equivalent to the double support phase in the Hybrid-LIP model. For more in-depth information,  one can refer to \cite{xiong_3-d_2022}.
\end{remark}

\subsection{Stabilization for Robot Dynamics}\label{sec::ro_stabilization}
Having formulated the S2S dynamics for the MLIP model as a discrete-time linear control system, we now aim to stabilize robot walking using it. In practice, we encounter a significant challenge in obtaining the S2S dynamics of the robot due to its inherently nonlinear nature. However, if we assume that the robot's control scheme ensures the availability of the next step, we can mathematically express its S2S dynamics as the Poincar\'{e} map:
\begin{align}
    \x_{k+1}^- = \Poin_{\x} (\x_k^-,\tau(t)).
\end{align}
Here, $\x_k^- \in \mathcal{TQ}$ is the robot state at the end of the UA phase, i.e., pre-impact states. The associated Poincaré map is denoted as $\Poin_{\x}$. Our focus lies on the pre-impact CoM states, which can be denoted $\x^R = [p^{\text{R},-}, L^{\text{R},-}]^T \in \R^2$. We can express the evolution of these pre-impact CoM states as 
\begin{align}
    \x_{k+1}^{\text{R}} = \Poin_{\x^\text{R}} (\x_k^{\text{R}},\tau(t)).
\end{align}

Using the MLIP S2S dynamics as an approximation, the robot S2S dynamics can be expressed as:
\begin{align}\label{eq::HLIPdynamicsApproximation}
    \x_{k+1}^{\text{R}} = A^\text{M} \x_k^{\text{R}} + B^\text{M} u_k^\text{R} +C^\text{M} + w, 
\end{align}
where $u_k^\text{R}$ denotes the $k$-th step size of the robot. $w = \Poin_{\x^\text{R}} (\x_k^{\text{R}},\tau(t)) -A^\text{M} \x_k^{\text{R}} -B^\text{M} u_k^\text{R} - C^\text{M}$ represents the model discrepancy, i.e. the integrated dynamics difference between the robot and the MLIP over a step. It is assumed that the realizable set of walking behaviors satisfies $w \in \mathbf{W}$, where $\mathbf{W}$ is a bounded set as mentioned in \cite{xiong_3-d_2022}. Let $\e:= \x^\text{R} - \x^\text{M}$ represent the error state, a stabilizing controller can be designed as follows:
\begin{align}\label{eq::HLIPcontroller}
    u^\text{R}_k = u^\text{M}_k + K (\x^\text{R}_k - \x^\text{M}_k),
\end{align}
which yields the error dynamics: 
\begin{align}
    \e_{k+1} = (A^\text{M} +B^\text{M} K)\e_{k} + w,
\end{align}
where $K$ is the controller gain. From linear control theory, any selections of $K$ that result in stable $A^\text{M}+B^\text{M} K$ can drive $\boldsymbol{e}$ to converge to an  invariant set $\mathbf{E}$ as in \cite{kouramas_minimal_2005}, i.e. if $\e_k \in \mathbf{E}$, $\e_{k+1} \in \mathbf{E}$. In this work, the controller gain $K$ was determined using the linear quadratic regulator. It's important to note that this process is independent of the specific model, as long as it can be represented as a linear discrete-time system. Similar techniques have been successfully employed in related works, such as \cite{kim_once-per-step_2017} for S2S dynamics derived from numerically linearized full robot dynamics and in \cite{xiong_3-d_2022} for a different reduced-order model.

\subsection{MLIP Periodic Orbits}
\begin{figure}[t]
    \centering
    \includegraphics[width = 1 \linewidth]{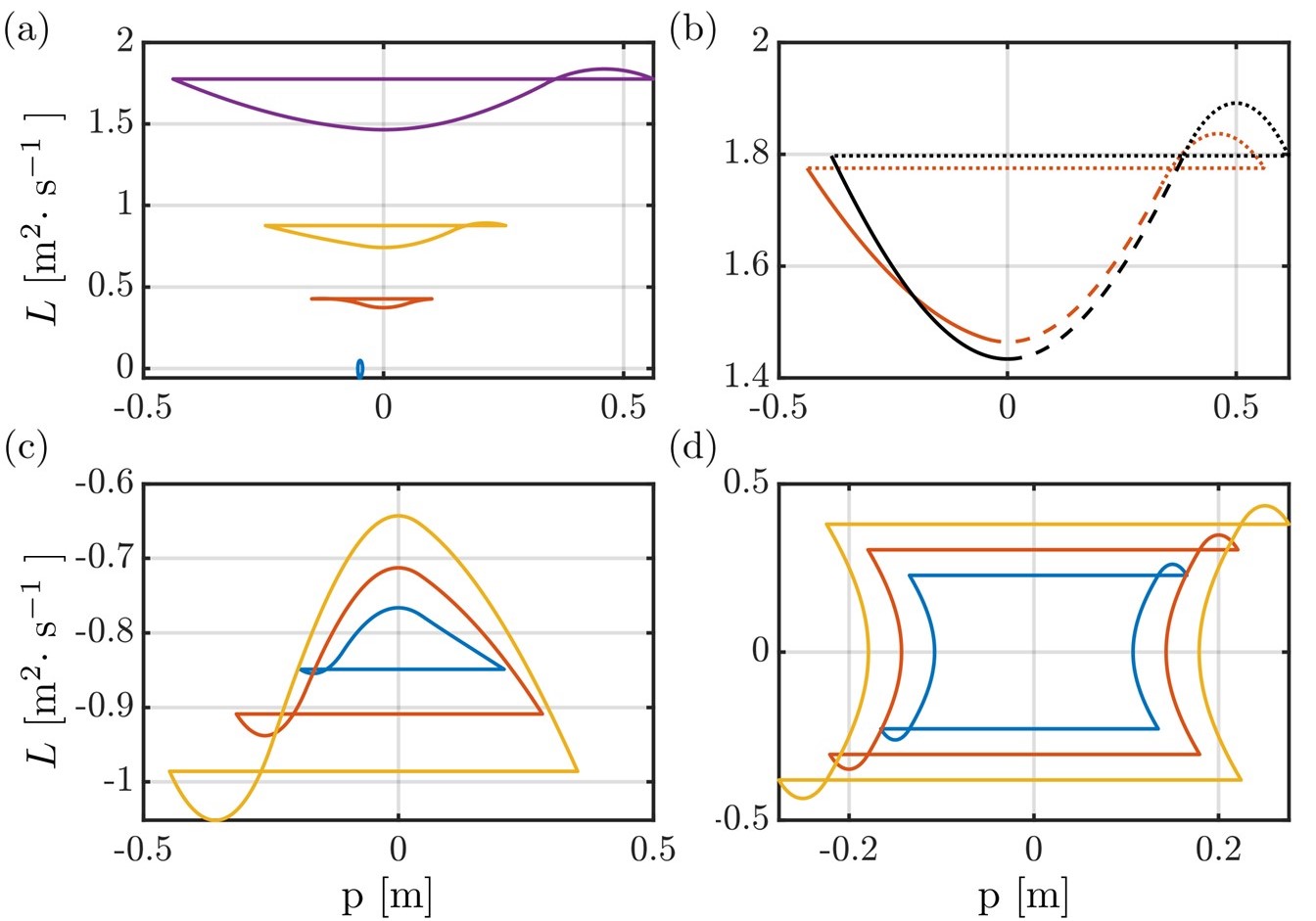}
    \vspace{-6mm}
    \caption{Phase portraits depicting various periodic orbits with $z_0 = 0.8$m: (a) Heel-to-toe walking at speeds of 0, 0.5, 1, 2 m/s shown in blue, red, yellow, and purple lines. (b) Comparison of heel-to-toe walking (red) and flat-footed walking (black) at 2m/s, highlighting FA, UA, and OA phases with solid, dashed, and dotted lines, respectively. (c) Toe-to-heel walking at -1 m/s, showcasing different total step times (T = 0.4s, 0.6s, and 0.8s) using blue, red, and yellow lines. (d) Flat-footed P2 orbit at 0m/s with nominal step widths of 0.3m, 0.4m, and 0.5m in blue, red, and yellow lines.}
    \vspace{-4mm}
    \label{fig::mlip_phase_portrait}
\end{figure}

In the context of walking, the robot is often required to follow a desired velocity. A straightforward approach is to employ closed-form reduced-order model periodic orbits, i.e. set $\x^\text{M} = \x^*$ to be the desired state. As MLIP is a planar model, it allows for decoupled planning of sagittal and lateral motion. We thus present the results for both Period-1 and Period-2 orbit that are suitable for sagittal and lateral planning, respectively. A visualization of the periodic orbits using different parameters is shown in Fig. \ref{fig::mlip_phase_portrait}.

\noindent \textbf{Period-1 Orbit:} The desired step size $u^*$ for a Period-1 orbit is determined by the desired walking velocity $v^d$ and the step duration $T$, where $u^* = v^d T$. The corresponding desired periodic pre-impact state for achieving $u^*$ is calculated by setting $\x_{k+1} = \x_k = \x^*$ in Eq. \eqref{eq::MLIP_S2S}:
\begin{align}\label{eq::P1xstar}
     \x^* = (I_{2 \times 2} - A^\text{M})^{-1}(B^\text{M} u^* + C^\text{M}),
\end{align}
where $I_{2 \times 2} \in \R^{2 \times 2}$ is the identity matrix. In this context, the controller can be expressed as:
\begin{align}\label{eq::MLIPcontroller}
    u^\text{R} = u^* + K (\x^\text{R} - \x^*). 
\end{align}

\noindent \textbf{Period-2 Orbit:} Unlike P1 orbits, there is no unique solution for the P2 orbit that achieves a desired velocity $v^d$. We use subscripts ${\text{L/}\text{R}}$ to denote the left or right stance leg. The step sizes must satisfy the equation $u_\text{L}^* + u^*_\text{R} = 2 v^d T$, and the choice of one step size determines the orbit. Solving $\x_{k+2} = \x_k$ yields the desired periodic pre-impact states:
\begin{align*}\label{eq::P2xstar}
    \x_\text{L/R}^* = (I_{2 \times 2}- (A^\text{M})^2)^{-1}( &A^\text{M} B^\text{M} u_\text{L/R}^* + B^\text{M} u^*_\text{R/L} \\ &+ A^\text{M} C^\text{M}+ C^\text{M}) . 
\end{align*}
Consequently, the controller can be written as $u^\text{R}_\text{L/R} = u^*_\text{L/R} + K (\x^\text{R} - \x^*_\text{L/R})$. When applying P2 orbits in the lateral plane, the state vector $\x$ is defined as $\x := [p_y, -L_x]^T$ to ensure consistency with the sign conventions used in the $A_\text{ct}$ matrix.

%% file: fullmodel.tex
\section{Robot Implementation}\label{sec::full_model}

The reduced-order model generates discrete commands using step-to-step stabilization. However, to translate these commands into real-time control signals for the physical robot, it is imperative to construct continuous control outputs and design corresponding feedback controllers. This process is detailed in this section.

\subsection{Output Definition}
\begin{figure}[t]
    \centering
    \includegraphics[width = 1 \linewidth]{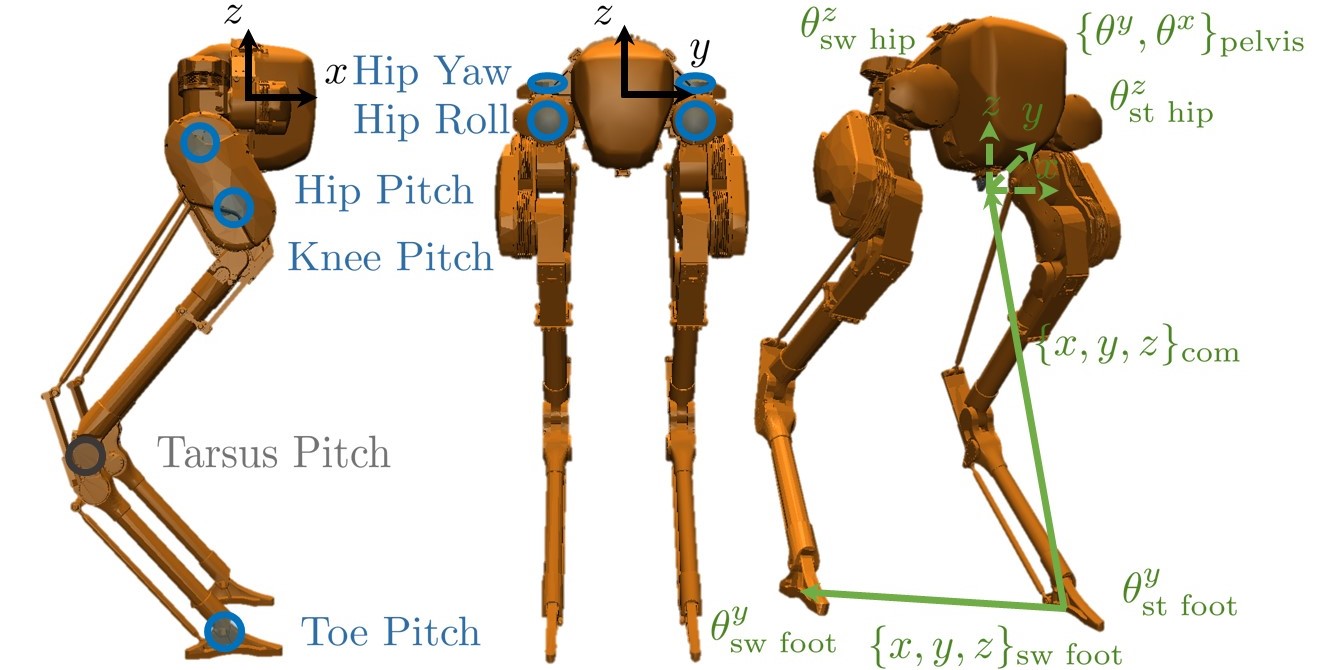}
    \vspace{-6mm}
    \caption{Robot Cassie schematics and output definition.}
    \vspace{-4mm}
    \label{fig::cassie_output}
\end{figure}
We have implemented our proposed method on robot Cassie, a 3D underactuated robot developed by Agility Robotics \cite{cassie}. As depicted in Fig. \ref{fig::cassie_output}, each leg of Cassie is modeled with 6 degrees of freedom (DOF), comprising five motored joints and one passive tarsus joint. When combined with the 6 DOF for the floating-base pelvis frame, the total DOF for the robot is 18.

To realize the MLIP-based approach on Cassie, the output design adheres to specific requirements. The vertical CoM position $z_\text{com}$ relative to the stance pivot should remain approximately constant. The vertical position of the swing foot $z_\text{st foot}$ is constructed to periodically lift off and strike the ground. The horizontal position of the swing foot $\{ x, y\}_\text{sw foot}$ relative to the stance pivot is controlled to achieve the desired step size $\{u_x, u_y\}$ from the MLIP-based stepping controller Eq. \eqref{eq::MLIPcontroller}. The pitch angles of the stance and swing foot should be controlled to provide the desired contact location corresponding to the desired walking mode. Additionally, the pelvis roll and pitch angles $\{ \theta^y, \theta^x\}_\text{pelvis}$ and the stance and swing hip yaw angles $\theta^z_\text{st hip}$, $\theta^z_\text{sw hip}$ should be controlled to fully constrain the walking behaviors. With these considerations in mind, the desired walking behavior is encoded by the virtual constraints \cite{grizzle_models_2014}, defined as:
\begin{align*}
    \Y = \h^{a} - \h^{d} \in \mathbb{R}^{12} ,
\end{align*}
where $\h^{a}$ and $\h^{d}$ denote the actual and desired output defined as follows:
\begin{equation*}
\begin{split}
    \h = \textrm{col}(& \{ x, y, z\}_\text{com}, \theta^z_\text{st hip}, \theta^y_\text{st foot}, \{ \theta^y, \theta^x\}_\text{pelvis},\\
        &\{ x, y, z\}_\text{sw foot}, \theta^z_\text{sw hip}, \theta^y_\text{sw foot}) \in \mathbb{R}^{12}.
        \end{split}
\end{equation*}

In Fig. \ref{fig::cassie_output}, we provide a visualization of these outputs, specifically for flat-footed walking. For heel-to-toe walking, the position of CoM is defined to be relative to the stance toe, and the position of the swing foot is represented as the vector from the stance toe to the swing heel. Similar modifications need to be applied to toe-to-heel walking.

Cassie employs a line foot design, which means that only the toe pitch motor is present on the foot link. Consequently, the robot lacks motor control for adjusting the foot roll angle. This limitation leads to inherent underactuation in the lateral plane during single support phases. Similarly, in the OA domain, where only the back toe and front heel make contact with the ground for heel-to-toe walking, the ZMP can solely reside along the line connecting these two contact points. This configuration results in coupled control of the xCoM and yCoM components. Therefore, achieving independent control of both xCoM and yCoM is not feasible during OA phases. These hardware constraints lead to the following choice of selection matrices $S_\ti$ for each domain such that $\h_{{\tFA,\tUA,\tOA}} = S_{{\tFA,\tUA,\tOA}} \h$:
\begin{subequations}
\begin{align}
    S_{\tFA} &= \textrm{diag}(\{1,0,1\}, 1,0, \{1,1\}, \{1,1,1\},1,1),\\
    S_{\tUA} &= \textrm{diag}(\{0,0,1\}, 1,1, \{1,1\}, \{1,1,1\},1,1),\\
    S_{\tOA} &= \textrm{diag}(\{0,0,1\}, 1,1, \{1,1\}, \{0,0,0\},0,1).
\end{align}
\end{subequations}

The selection matrix for each domain aligns with the holonomic constraints present in that specific domain. In the UA and OA phases, we assume patch contact with the ground, meaning that the contact points' positions {x, y, z} and yaw angle are constrained. During the FA phase, we apply the line contact assumption, which adds the constraint on the contact pitch angle. In the current formulation, we allow the horizontal CoM states to evolve passively during the OA phase. This choice helps avoid mismatched commands for xCoM and yCoM control during that phase.


The construction of the most desired outputs for our control framework is consistent with the approach outlined in \cite{xiong_3-d_2022}. However, there are exceptions in the case of the horizontal COM position, stance foot pitch, and swing foot pitch angles. 


\subsubsection{Horizontal COM position}

During the FA phase, one option is to control the ZMP to transit from the heel to the toe, as assumed in the MLIP model construction. However, this approach requires feedback control of the ZMP position at the jerk level, which is impractical. Instead, we opt to directly regulate the horizontal COM states at the end of the FA phase to reach a desired state denoted as $\x^{\text{M},*}_{\text{FA}^-}$. This state can be calculated using the MLIP step-to-step dynamics, with the Poincaré section instead chosen to be the end of the FA phase, i.e., FA$^-$. The trajectory for this control is defined using a Bézier polynomial that can be written as:
\begin{align}
    x_\text{com}^d(s_\tFA) \coloneqq b_\text{xcom}(s_\tFA) = A(s_\tFA) \alpha_\text{xcom},
\end{align}
where $s_\tFA \coloneqq \frac{t_\tFA}{T_\tFA} \in [0, 1)$ is the phase variable in FA domain, and $A(s_\tFA) \in \R^{1 \times n_b}$ is defined using the definition of Bézier polynomial. The $k$-th element of $A(s_\tFA)$ is computed as:
\[ A_k(s_\tFA) = \frac{n_b!}{k!(n_b-k)!}s_\tFA^k (1-s_\tFA)^{n_b-k}.\]
Additionally, $\alpha_\text{xcom} \in \R^{n_b}$ represents the coefficients of a Bézier polynomial of degree $n_b$. These coefficients are determined at the start of the FA phase, subject to the following linear equality constraints:
\begin{align}
    \begin{bmatrix}
        A(0) \\ \dot{A}(0,T_\text{FA}) \\ A(1) \\ \dot{A}(1,T_\text{FA}) 
    \end{bmatrix} \alpha_\text{xcom} = \begin{bmatrix}
        x_\text{com, FA+}^a \\ \dot{x}_\text{com, FA+}^a \\ p^{\text{M},*}_{\text{FA}^-} \\ \frac{1}{z_0} L^{\text{M},*}_{\text{FA}^-} 
    \end{bmatrix} .
\end{align}
Here, $\dot{A}(s_\tFA,T_\text{FA}) \in \R^{1 \times n_b}$ is also determined based on the definition of the Bézier polynomial, ensuring that $\dot{b}_\text{xcom}(s_\tFA,T_\text{FA}) = \dot{A}(s_\tFA,T_\text{FA}) \alpha_\text{xcom}$. The actual horizontal CoM position and velocity at the beginning of the domain is given by $x_\text{com, FA+}^a$ and  $\dot{x}_\text{com, FA+}^a$. Notably, the second coordinate in $\x^{\text{M},*}_{\text{FA}^-}$ requires conversion from angular momentum to velocity, as indicated.

\subsubsection{Stance and swing foot pitch angle}
The desired end-of-domain stance and swing foot pitch angles vary for different contact modes in different domains. For instance, in the UA phase of heel-to-toe walking, the final stance foot angle should be greater than zero to enable heel-lift motion. However, these angles share a common output structure:
\begin{align}
    \{\theta^y_\text{sw/st foot}\}^d(s_\ti)& \coloneqq (1-b_\text{foot}(s_\ti)) \{\theta^y_\text{sw/st foot}\}^a_\text{i+} \\& + b_\text{foot}(s_\ti) \{\theta^y_\text{sw/st foot}\}^*_\ti, \nonumber
\end{align}
where $s_\ti \coloneqq \frac{t_\ti}{T_\ti} \in [0, 1)$ is the phase variable within each domain, $\{\theta^y_\text{sw/st foot}\}^a_\text{i+}$ is the actual foot pitch angle at beginning of i-th domain, $\{\theta^y_\text{sw/st foot}\}^*_\ti$ denotes the desired end-of-domain pitch angle, $b_\text{foot}(s_\ti)$ represents a Bézier polynomial that transitions from 0 to 1.

\subsection{Feedback Controller} 
Using the synthesized outputs, we employ a task-space quadratic programming (QP) based controller \cite{bouyarmane_quadratic_2019} to ensure the tracking of desired trajectories while respecting the constrained dynamics, physical motor torque limits, and ground contact forces constraints. In each domain i and at each control loop, we formulate the QP with optimization variables $\boldsymbol{\ddot{q}}, \boldsymbol{\tau}, \boldsymbol{f}_\ti$ as follows: 
\begin{align}
   \underset{\boldsymbol{\ddot{q}}, \boldsymbol{\tau}, \boldsymbol{f}_\ti } {\text{min}}  & \quad ||\ddot{\h}^a_\ti(q,\dot{q},\ddot{q}) - \ddot{\h}^d_\ti - \ddot{\Y}^t_\ti ||^2_Q, \label{eq::TSC} \tag{TSC-QP} \\
\text{s.t.}  & \quad   \text{Eq.}~\eqref{eq::eom}, \eqref{eq::hol},  \tag{Dynamics} \\
 & \quad    A_{\text{GRF}} \boldsymbol{f}_\ti  \leq \boldsymbol{b}_{\text{GRF}}, \tag{Contact} \\ 
 & \quad  \boldsymbol{\tau}_{lb} \leq \boldsymbol{\tau} \leq \boldsymbol{\tau}_{ub}.  \tag{Torque Limit}
\end{align}
Here, $Q$ denotes a weight matrix, and $ \ddot{\Y}^t = - K_p \Y - K_d \dot{\Y}$ is the target acceleration of the output that enables exponential tracking, where $K_p, K_d$ are the proportional and derivative gains. 
The affine contact constraint on $\boldsymbol{f}_\ti$ approximates the contact friction cone constraint. $\boldsymbol{\tau}_{lb}$ and $\boldsymbol{\tau}_{ub}$ represent the lower and upper torque bounds. Solving this QP yields the optimal torque $\boldsymbol{\tau}$ that is applied to the robot. 

%% file: results.tex


\section{Results}\label{sec::results}
We evaluate the proposed approach using our \texttt{C++} implementation in the open-sourced simulator \cite{cassie_mujocosim} with the Mujoco physics engine \cite{todorov_mujoco_2012} on the robot Cassie. The output construction and corresponding low-level controller, as described in \eqref{eq::TSC}, are executed at a rate of 1kHz for real-time control. A visual representation of the results is available in the supplementary video provided earlier.

For our MLIP planning, we assume a constant CoM height of $z_0 = 0.8$m and a foot length of $\rho = 0.16$m, consistent with Cassie's physical foot length. Since lateral dynamics are always underactuated during the single support phase, we use $T_\text{SS}$ to denote the step duration for the single support phase, where $T_\text{SS} = T_\tFA + T_\tUA$ for both the sagittal and lateral planes.  In all tests, we use $T_\text{SS} = 0.4$s and $T_\text{OA} = 0.1$s. For the MLIP model in the sagittal plane, we set $T_\tFA = T_\tUA = \frac{T_\text{SS}}{2}$, resulting in a phase distribution of 40\% FA, 40\% UA, and 20\% OA, which aligns with actual human walking data \cite{birch_terminology_2015}. In the lateral plane, we have $T_\tFA = 0$ and $ T_\tUA = T_\text{SS}$. 

\begin{figure}[t]
\includegraphics[width=\linewidth]{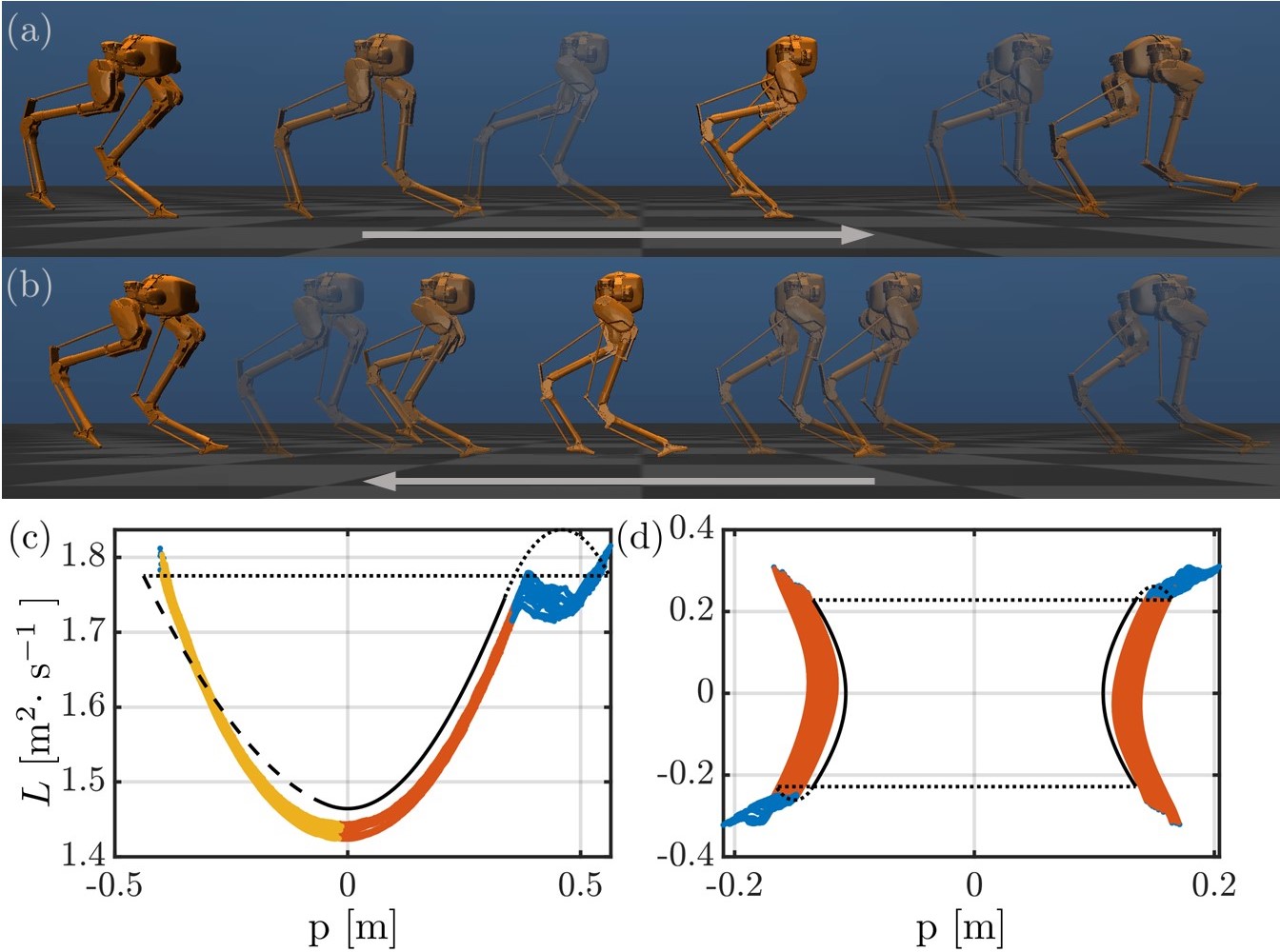}
 \vspace{-6mm}
  \caption{(a) Heel-to-toe walking at a speed of 2 m/s. (b) Toe-to-heel walking at -1.5m/s. (c) Phase portrait depicting 2m/s steady-state heel-to-toe walking. The yellow, red, and blue lines represent the FA, UA, and OA phases for the robot, following the color code from Fig. \ref{fig::lip_h2t_defintion}. The dashed, solid, and dotted lines correspond to the FA, UA, and OA phases for the periodic orbit in the MLIP model. (d) Phase portrait illustrating steady-state lateral walking, also using the same color code.}
  \vspace{-4mm}
  \label{fig::cassie_walking_results}
\end{figure}
\begin{figure}[t]
  \includegraphics[width=\linewidth]{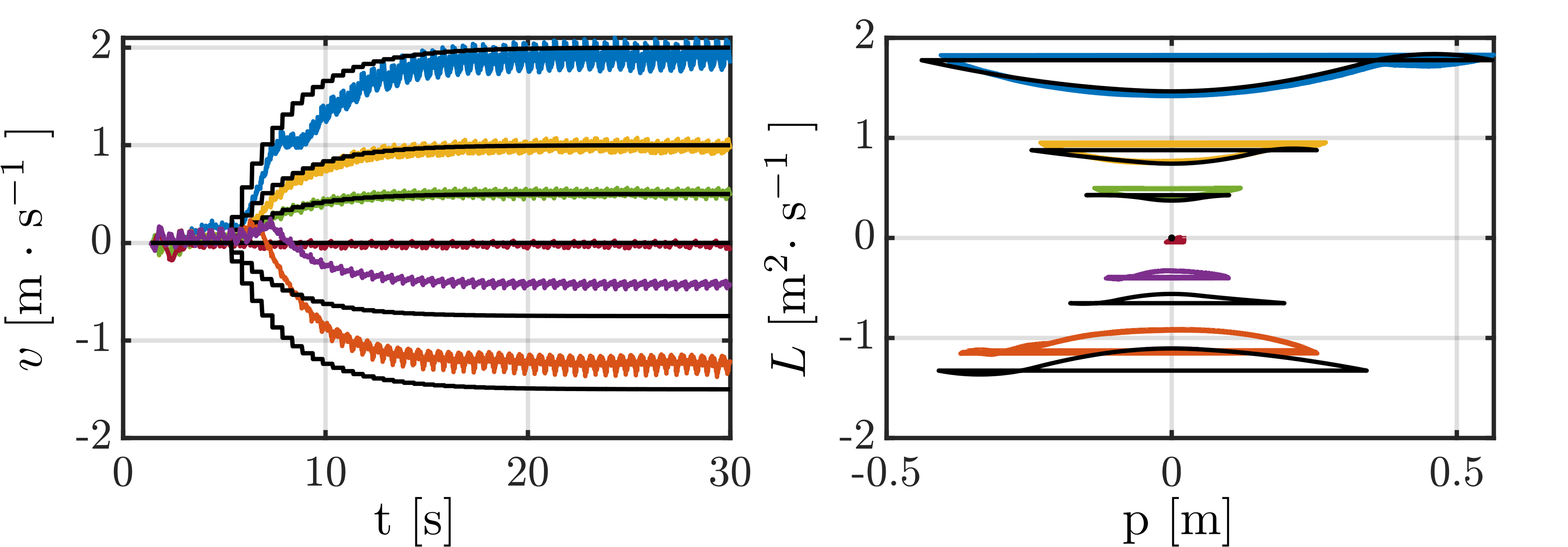}
  \vspace{-4mm}
  \caption{Velocity tracking performance and corresponding steady-state phase portrait for a range of walking speeds: 2 m/s, 1 m/s, 0.5 m/s, 0 m/s, -0.75 m/s, and -1.5 m/s, depicted by blue, yellow, green, burgundy, purple, and red lines, respectively. The commanded velocity is given in solid black lines. }
  \vspace{-6mm}
  \label{fig:speed_comparison}
\end{figure}

\subsection{Versatile Walking}

The versatility of the proposed approach enables us to showcase a wide range of walking behaviors using Cassie. In all test scenarios, the robot initiates from a stationary posture and is initially commanded to step-in-place for a duration of five seconds. Subsequently, the reference velocity gradually changes to the desired value.

Fig. \ref{fig::cassie_walking_results}(a) and (b) provide a glimpse of the robot's steady-state walking under speeds of 2 m/s and -1.5 m/s, respectively. Employing the multi-domain formulation, Cassie exhibits a natural walking motion characterized by human-like foot rolls with distinct UA, OA, and FA phases. In Fig. \ref{fig::cassie_walking_results}(c), we offer a comparison of the phase portrait for the periodic orbit realized by Cassie and the periodic orbit defined for MLIP for 2m/s heel-to-toe walking. Fig. \ref{fig::cassie_walking_results}(d) depicts the phase portrait for lateral walking using a P2 orbit.

In Fig. \ref{fig:speed_comparison}, we examine the performance of our approach across a range of reference velocities. Remarkably, at all reference velocities, the walking stabilizes near the reference velocity, and the robot's states remain within a small, bounded error set relative to the nominal MLIP states as shown in the phase portrait.

\begin{remark}
Our primary objective in this work is not to achieve perfect velocity tracking performance. As discussed in Sec. \ref{sec::ro_stabilization}, our proposed reduced-order model planner is designed to stabilize the robot within a small bounded error set around the planned reduced-order model trajectory. To improve the global tracking performance, there are two viable strategies. One approach involves integrating a high-level planner to adjust the desired velocity sent to the robot. Alternatively, data-driven techniques \cite{dai_data-driven_2023} can be employed to reduce robot-dependent model discrepancies, thereby reducing the size of the error set. 
\end{remark}

\subsection{Push Recovery}
\begin{figure}[t]
  \includegraphics[width=\linewidth]{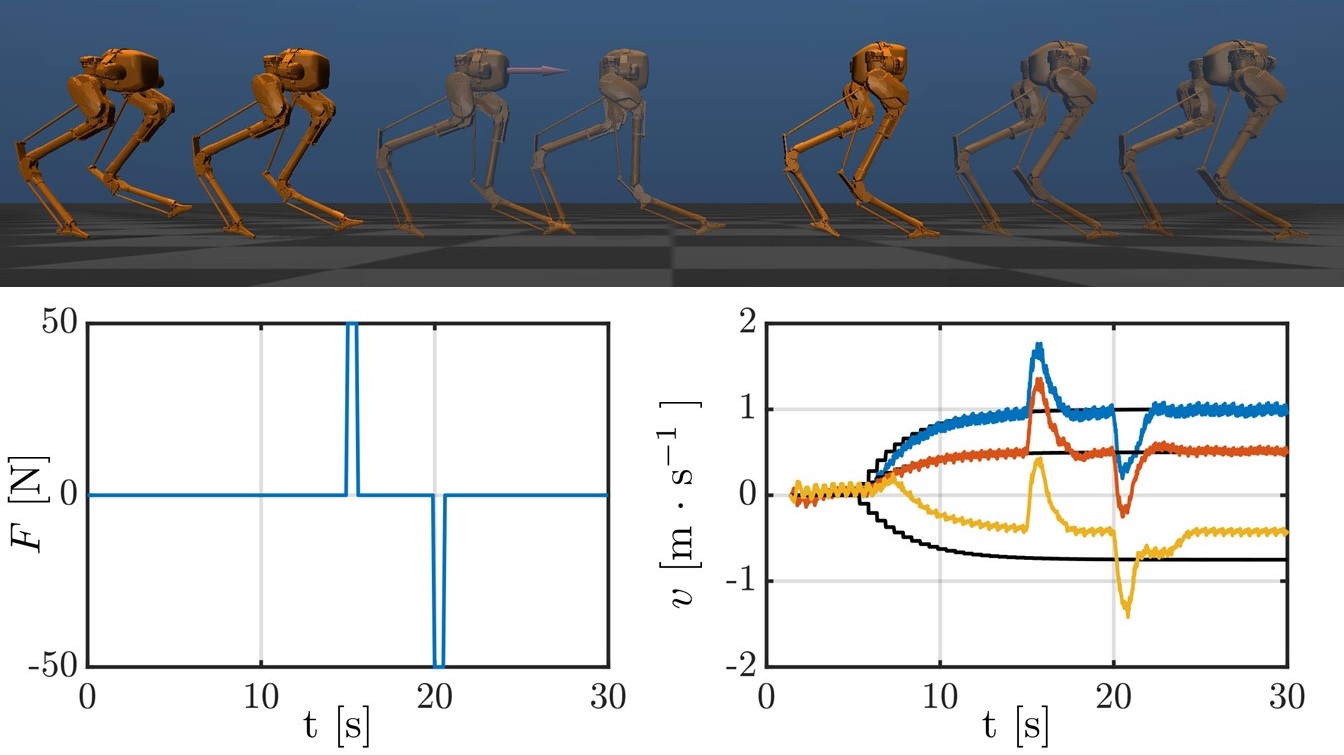}
  \vspace{-6mm}
  \caption{(top) Cassie's recovery from a 50N push during 1m/s walking. (bottom) Disturbance force profile and corresponding velocity tracking results for push recovery tests at walking speeds of 1m/s, 0.5m/s, and 0.75m/s, represented by blue, red, and yellow lines respectively.}
  \vspace{-4mm}
  \label{fig::push_recovery}
\end{figure}

The proposed method uses online foot placement planning to stabilize the robot. To evaluate the robustness of this approach, we apply unknown disturbance forces to Cassie. Specifically, as shown in Fig. \ref{fig::push_recovery}(b), we applied a 50N and -50N push to the pelvis of the robot at times 15s and 20s, where each lasts for 0.5s.

Fig. \ref{fig::push_recovery}(a) shows the robot's response to the 50N push during heel-to-toe walking at a speed of 1m/s. Evidently, the robot quickly recovers from the perturbation by taking a few larger steps. Fig. \ref{fig::push_recovery} (c) displays the CoM velocity profiles for three different commanded speeds when subjected to the same disturbance.  Notably, all commanded speeds exhibit the ability to withstand external forces and subsequently resume normal walking behavior.

\subsection{Maximum Speed}
\begin{figure}[t]
  \includegraphics[width=\linewidth]{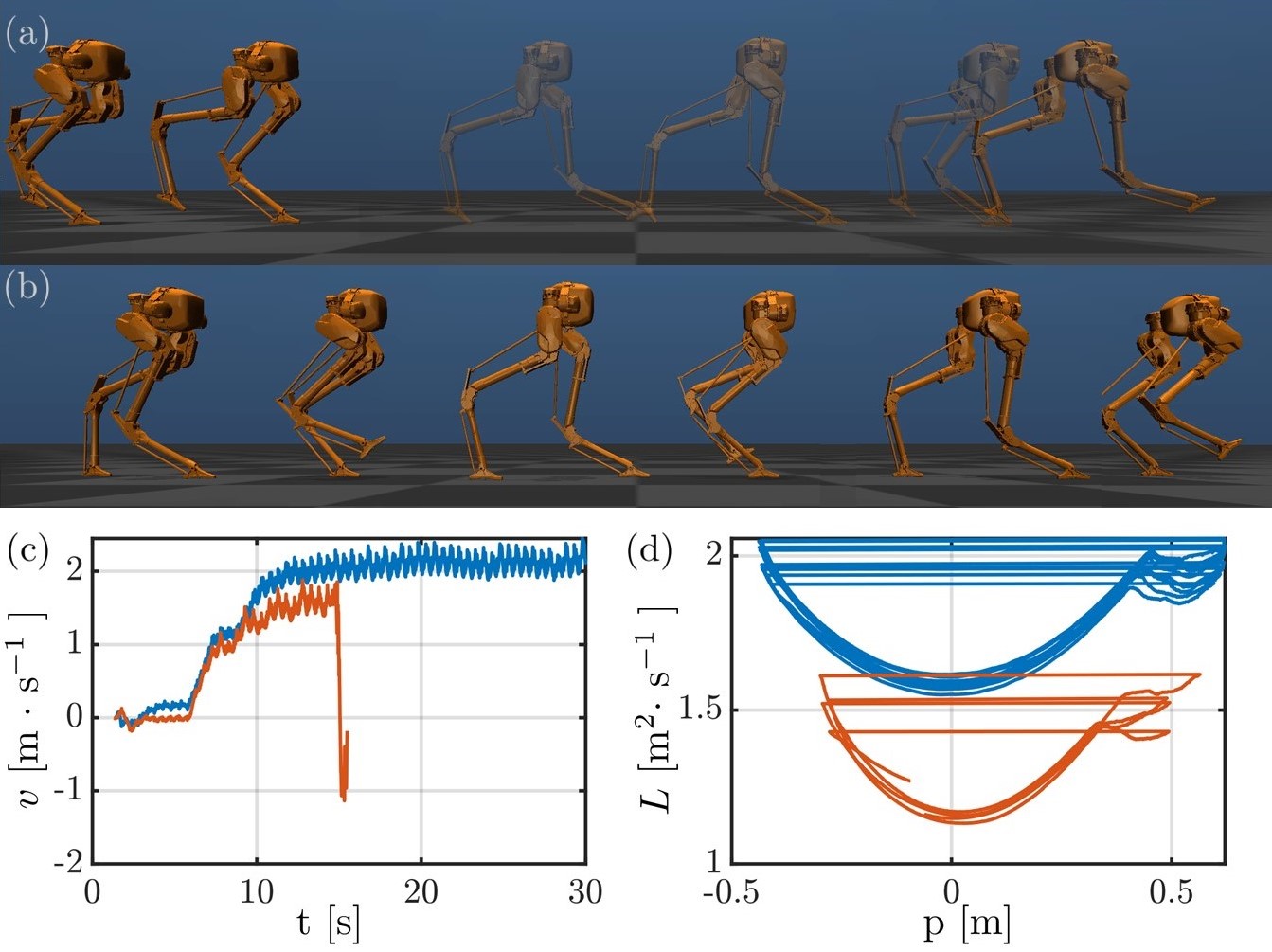}
  \vspace{-6mm}
  \caption{(a\&b) Cassie's walking behavior under maximum attainable speed for heel-to-toe walking and flat-footed walking. (c\&d) Velocity profiles and phase portraits for the two scenarios, where blue and red lines are used for heel-to-toe walking and flat-footed walking respectively. For flat-footed walking, the robot oscillates around its steady-state orbits and eventually falls. }
  \vspace{-4mm}
  \label{fig::maximum_speed}
\end{figure}
Multi-domain walking offers the potential for achieving larger footsteps compared to flat-footed walking, ultimately leading to higher walking speeds. Our results effectively showcase this advantage in Fig. \ref{fig::maximum_speed}, where we compare the walking behavior for the maximum attainable speed based on the specified gait parameters in (a) and (b). It is clear that heel-to-toe walking results in a noticeably larger footstep. The corresponding velocity profile and phase portrait are presented in Fig. \ref{fig::maximum_speed}(c) and (d). The proposed method using heel-to-toe walking achieves an impressive speed of approximately 2.15 m/s, whereas flat-footed walking only reaches a speed of 1.65 m/s. To provide context, the average human walking speed is 1.42 m/s. We are able to realize highly dynamic locomotion behaviors on Cassie using our proposed method. 

\begin{remark}
The previous method \cite{xiong_3-d_2022} was capable of achieving a maximum speed of approximately 2 m/s on Cassie. However, this was only attainable with much faster stepping motions, specifically with $T_\text{SS} = 0.3$s and $T_\text{OA} = 0$s. It's important to note that the resulting gait was unstable without additional data-driven adaptation, as discussed in \cite{dai_data-driven_2023}.
\end{remark}

\section{Conclusion}\label{sec::conclusion}
In conclusion, this paper introduces a novel reduced-order model based approach to realize multi-domain walking on bipedal robots. Leveraging the S2S dynamics of the proposed MLIP model, we have demonstrated the ability to stabilize the robot at arbitrary walking speed, achieving a remarkable maximum speed of 2.15 m/s. The robustness of the method is demonstrated through push recovery tests. Importantly, this method can be readily extended to applications in robotic assistive devices, such as exoskeletons, to enable human-like multi-domain locomotion for the mobility impaired. 